\documentclass[lettersize,journal]{IEEEtran}

\usepackage{amsmath,amsfonts}
\usepackage{algorithmic}
\usepackage{algorithm}
\usepackage{array}
\usepackage[caption=false,font=normalsize,labelfont=sf,textfont=sf]{subfig}
\usepackage{textcomp}
\usepackage{stfloats}
\usepackage{url}
\usepackage{verbatim}
\usepackage{graphicx}
\usepackage{cite}
\usepackage{booktabs}
\usepackage{multirow}
\usepackage{xkeyval}
\usepackage{wrapfig}
\usepackage{xparse}
\usepackage[colorlinks=true,linkcolor=red,citecolor=blue,urlcolor=black,pdfborder={0 0 0}]{hyperref}

\usepackage[T1]{fontenc}

\makeatletter
\newwrite\authorbibfile
\AtBeginDocument{%
  \immediate\openout\authorbibfile=\jobname.aub%
}
\AtEndDocument{%
  \immediate\closeout\authorbibfile
  \InputIfFileExists{\jobname.aub}{}{}
}

\define@key{authorbib}{scale}[1]{\def\AuthorbibKVMacroScale{#1}}
\define@key{authorbib}{wraplines}[10]{\def\AuthorbibKVMacroWraplines{#1}}
\define@key{authorbib}{imagewidth}[4cm]{\def\AuthorbibKVMacroImagewidth{#1}}
\define@key{authorbib}{overhang}[10pt]{\def\AuthorbibKVMacroOverhang{#1}}
\define@key{authorbib}{imagepos}[l]{\def\AuthorbibKVMacroImagepos{#1}}
\makeatother

\presetkeys{authorbib}{imagepos=l,imagewidth=4cm,wraplines=8,overhang=20pt}{}

\newlength{\AuthorbibTopSkip}
\newlength{\AuthorbibBottomSkip}
\NewDocumentCommand{\authorbibliography}{O{} m m m}{%
  \setkeys{authorbib}{#1}%
  \immediate\write\authorbibfile{%
    \string\begin{wrapfigure}[\AuthorbibKVMacroWraplines]{\AuthorbibKVMacroImagepos}[\AuthorbibKVMacroOverhang]{\AuthorbibKVMacroImagewidth}^^J
    \string\includegraphics[scale=\AuthorbibKVMacroScale]{#2}^^J
    \string\end{wrapfigure}^^J
    \unexpanded{\vspace{\AuthorbibTopSkip}}^^J
    \string\noindent\relax
    \unexpanded{\textbf{#3}\par}^^J
    \string\noindent\relax
    \unexpanded{#4}^^J
    \unexpanded{\vspace{\AuthorbibBottomSkip}}^^J
  }%
}

\begin{document}

\title{MVFA: A Multi-View
Text-Guided Multimodal Fusion LLM Adapter for Sentiment Analysis and Emotion Recognition}

\author{%
	Pengfei Shao,
	Jisheng Dang,
      Jiawen Fang,
      Ning Liu,
      Wencan Zhang,
            Bimei Wang,
            Jingwen Zhao,
    	Jianhuang Lai, 
 ~\IEEEmembership{Senior Member, ~IEEE},
    	Qi Tian,~\IEEEmembership{Fellow, ~IEEE},
	Tat-Seng Chua%
    \thanks{This work was supported by the National Natural Science Foundation of China (Grants No. 62227807 and U24B20186). This work was also supported by the Supercomputing Center of Lanzhou University.}
    \thanks{Pengfei Shao and Ning Liu are with the School of Information Science and Technology, Guangdong University of Foreign Studies, Guangdong, China (E-mail: 20251010019@mail.gdufs.edu.cn).}
    \thanks{Jisheng Dang, Bimei Wang, and Jingwen Zhao are with the School of Information Science and Engineering, Lanzhou University, Lanzhou, China (E-mail: dangjisheng@lzu.edu.cn).}
    \thanks{Jiawen Fang and Jianhuang Lai are with the School of Computer Science and Engineering, Sun Yat-sen University, Guangdong, China (E-mail: fangjw6@mail2.sysu.edu.cn).}
 \thanks{Qi Tian is with Cloud and AI BU, Huawei, Shenzhen, Guangdong 518129,
China (E-mail: tian.qi1@huawei.com).}
    \thanks{Wencan Zhang and Tat-Seng Chua are with the School of Computing, National University of Singapore, Singapore 119077 (E-mail: dcscts@nus.edu.sg). }
    \thanks{*Corresponding author: Jisheng Dang.}
}

\maketitle

\begin{abstract}
Multimodal sentiment analysis and emotion recognition in conversations demand effective modeling of heterogeneous interactions across textual, acoustic, and visual modalities. Although large language models (LLMs) offer powerful language understanding, adapting them to multimodal affective computing remains challenging: full-model fine-tuning is computationally prohibitive, while many existing lightweight adapters fail to preserve rich textual cues during cross-modal fusion. To address these limitations, we propose the multi-view text-guided multimodal fusion adapter (MVFA), a parameter-efficient framework that augments frozen LLMs with strong multimodal reasoning capability. MVFA first constructs complementary text views via max pooling, mean pooling, and attention pooling; these views then guide cross-modal interactions with audio and visual features. The fused multimodal representations are subsequently compressed into a compact set of learnable pseudo-tokens through an Enhanced Q-Former Fusion Module. Using ChatGLM3-6B-base as the primary backbone, we further validate MVFA on LLaMA2-7B and Qwen3-8B to examine its portability across multiple frozen LLM backbones. MVFA is evaluated on three challenging datasets: CH-SIMS V2.0, MELD, and CHERMA. Experimental results demonstrate that MVFA achieves state-of-the-art performance on key metrics while updating only a small fraction of parameters. Specifically, it attains 84.62\% Acc2 and 84.59\% F1 on CH-SIMS V2.0, 67.36\% Acc and 66.03\% WF1 on MELD, and 74.66\% Acc on CHERMA. These findings establish multi-view text-guided fusion as an effective and scalable paradigm for parameter-efficient multimodal LLM adaptation in affective computing. The code is publicly available at \url{https://github.com/Overwhelm1208/MVFA}.
\end{abstract}


\section{Introduction}
\begin{figure}[t]
    \centering
    \includegraphics[width=0.5\textwidth]{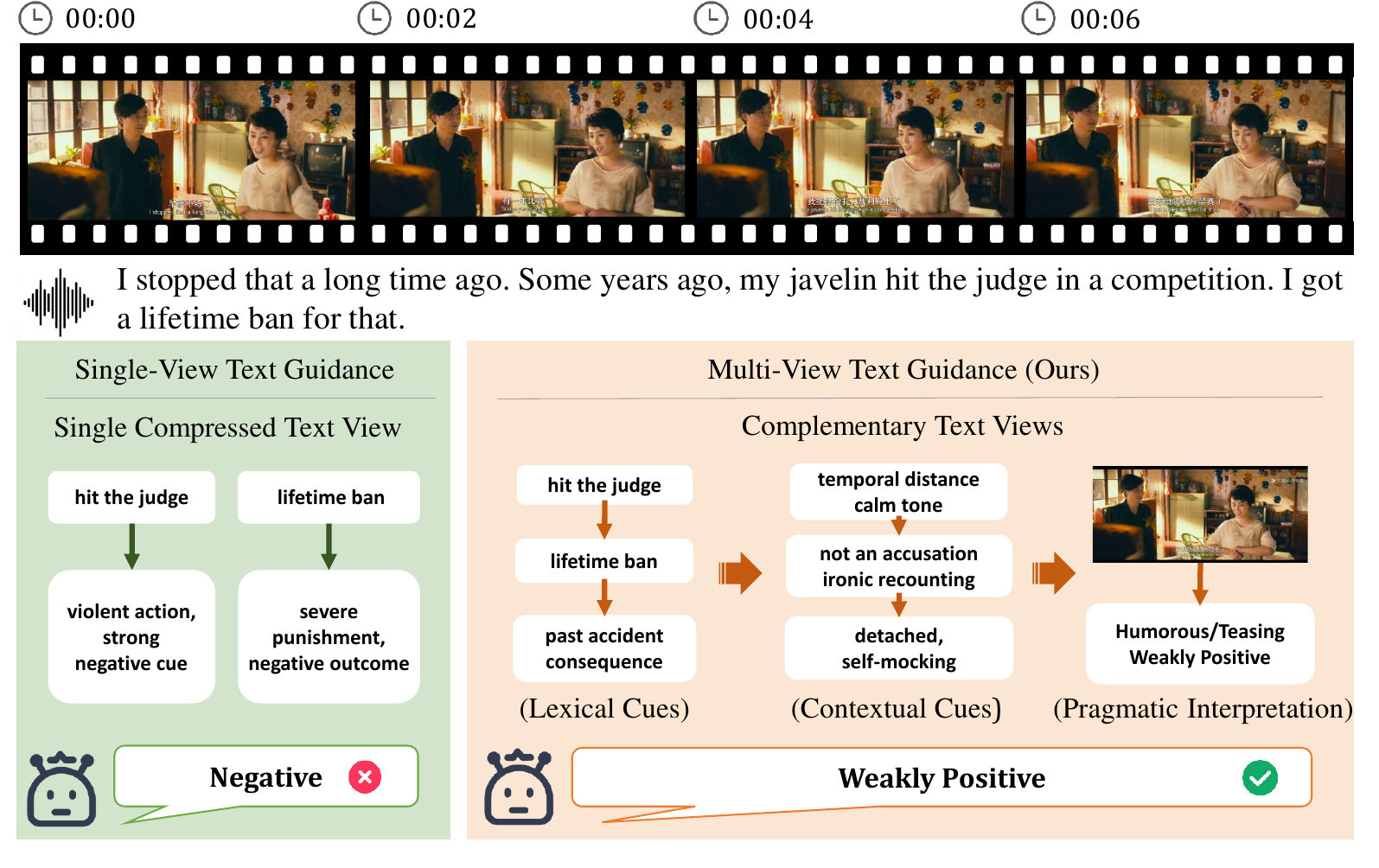}
    \caption{An example illustrating that a single compressed text
    view may fail to preserve complementary affective cues (e.g.,
    salient lexical expressions, discourse-level contrast, and
    context-relevant tokens) during multimodal fusion.}
    \label{fig:motivation}
\end{figure}

\IEEEPARstart{H}{uman} communication is inherently multimodal, with
intentions and emotions conveyed through the coordinated use of language, acoustic prosody, and visual cues. This observation has driven extensive research in multimodal sentiment analysis (MSA) and emotion recognition in conversations (ERC), which aim to infer speakers' affective states by integrating heterogeneous signals from
text, audio, and vision~\cite{poria2019meld,zadeh2018multimodal}. Such
capabilities are central to building empathetic dialogue systems and socially aware artificial intelligence.
Recent advances in large language models (LLMs) have opened new opportunities for multimodal affective computing. In particular, frozen LLMs provide powerful language understanding and generation capabilities, making them appealing backbones for parameter-efficient
adaptation. However, directly applying LLMs to MSA and ERC remains a challenge. Full fine-tuning of large multimodal systems is computationally prohibitive, while existing lightweight multimodal adapters typically rely on a single compressed textual view or dense cross-modal interactions, which often fail to preserve complementary textual cues essential for aligning language with nonverbal behaviors. As illustrated in Fig.~\ref{fig:motivation}, a single utterance may contain multiple affective cues simultaneously, such as salient sentiment-bearing words, discourse-level contrast, and context-relevant tokens that should be emphasized differently during multimodal fusion.

Earlier multimodal approaches primarily focused on explicit fusion of
heterogeneous representations. For instance, TFN models unimodal,
bimodal, and trimodal interactions via
tensor products~\cite{zadeh2017tensor}, while LMF improves efficiency
by factorizing high-dimensional fusion
tensors~\cite{liu2018efficient}. Subsequent methods introduced more
adaptive mechanisms, such as MAG that conditions pretrained language
representations on audio and visual
cues~\cite{rahman2020integrating}, CHFN that dynamically adjusts word
representations using nonverbal
information~\cite{guo2022dynamically}, and GraphCFC that models
cross-modal complementarity through graph-based message
passing~\cite{li2023graphcfc}. More recently, parameter-efficient
multimodal adaptation has gained increasing attention in the LLM era.
MSE-Adapter~\cite{yang2025mse}, for example, injects compact
pseudo-tokens into frozen LLMs for MSA and ERC. Nevertheless, existing
approaches still leave significant room for improvement in how
complementary textual cues are preserved and leveraged to guide
multimodal alignment before token injection.

To address these limitations, we propose the \textbf{Multi-View Text-Guided Multimodal Fusion Adapter (MVFA)}, a parameter-efficient framework that adapts frozen LLMs to MSA and ERC. Instead of representing text as a single compressed vector, MVFA constructs complementary text views through max pooling, mean pooling, and attention pooling; these views then guide cross-modal interactions with audio and visual features. The fused multimodal representations are subsequently compressed into a compact set of learnable pseudo-tokens via an Enhanced Q-Former Fusion Module (ENQF). This design avoids parallel modality-specific branches and heavy stacking of self-attention while preserving richer textual guidance for multimodal reasoning.

We primarily implement MVFA on top of the open-source ChatGLM3-6B-base backbone~\cite{glm2024chatglm} and evaluate it on three widely used public datasets: CH-SIMS V2.0~\cite{liu2022make}, MELD~\cite{poria2019meld}, and CHERMA~\cite{sun2023layer}. To examine whether MVFA can be applied to different frozen LLM backbones, we further instantiate MVFA on LLaMA2-7B~\cite{touvron2023llama} and Qwen3-8B~\cite{qwen3technicalreport} in the main benchmark comparisons. Experimental results demonstrate that MVFA
achieves strong and consistent performance while updating only a small
fraction of model parameters. Our main contributions are summarized as follows:
\begin{itemize}
    \item We propose MVFA, a lightweight parameter-efficient adapter
    that effectively adapts frozen LLMs to both MSA and ERC tasks.
    \item We design a multi-view text modeling module that constructs
    complementary textual representations via max, mean, and attention
    pooling, progressively fusing them to guide cross-modal alignment.
    \item We introduce an Enhanced Q-Former Fusion Module (ENQF) that
    compresses multimodal information into compact pseudo-tokens for
    efficient injection into frozen LLMs.
    \item We conduct extensive experiments on CH-SIMS V2.0, MELD, and CHERMA, demonstrating that MVFA delivers strong and consistent results in both Chinese and English multimodal affective computing datasets and shows portability to different frozen LLM backbones, while ChatGLM3-6B-base gives the strongest overall performance in our experiments.
\end{itemize}

\section{Related Work}
\subsection{Conventional Multimodal Fusion for MSA and ERC}
Early multimodal approaches mainly focused on explicit fusion of heterogeneous representations. TFN \cite{zadeh2017tensor} models unimodal, bimodal, and trimodal interactions through tensor products, while LMF \cite{liu2018efficient} improves computational efficiency by factorizing high-dimensional fusion tensors. Subsequent methods introduced more adaptive interaction mechanisms. MAG \cite{rahman2020integrating} conditions pretrained language representations on acoustic and visual signals, CHFN \cite{guo2022dynamically} dynamically adjusts word representations using nonverbal information, and GraphCFC \cite{li2023graphcfc} models cross-modal complementarity in conversational settings via graph-based message passing. Other studies explore auxiliary supervision, disentangled representations, or robustness under missing modalities, such as Self-MM \cite{yu2021learning}, MISA \cite{hazarika2020misa}, and ConFEDE \cite{yang2023confede}. These methods provide useful mechanisms for multimodal interaction, but most are not designed for parameter-efficient adaptation of frozen LLMs.

\subsection{Unified-task and Multimodal Reasoning Frameworks}
Recent work has also explored more unified formulations of multimodal affective computing. UniMSE \cite{hu2022unimse} studies MSA and ERC within a shared framework, while broader multimodal models such as UNIMO \cite{li2021unimo} investigate unified multimodal understanding and generation. These approaches highlight the benefit of sharing knowledge across tasks and modalities. However, their focus is different from ours: rather than unifying tasks under a shared framework, we study how to adapt a frozen LLM to multimodal sentiment and emotion reasoning in a lightweight and parameter-efficient manner.

\subsection{Parameter-Efficient Adaptation for Frozen LLMs}
With the rise of large pretrained models, parameter-efficient adaptation has become increasingly important. Prompt-based approaches such as CoOp \cite{zhou2022conditional} and MaPLe \cite{khattak2023maple} introduce learnable tokens to guide downstream behavior, while adapter-based methods insert lightweight trainable modules into frozen backbones. In multimodal settings, MMA \cite{yang2024mma} balances modality-specific adaptation and cross-modal alignment through shared and private projections, Chameleon \cite{liaqat2024chameleon} unifies heterogeneous modalities into RGB-like tokens for single-branch processing, and MSE-Adapter \cite{yang2025mse} injects compact pseudo-tokens into frozen LLMs for MSA and ERC. Among these methods, MSE-Adapter is the most closely related to our work because both methods follow the frozen-LLM plus pseudo-token adaptation paradigm. The key difference is that MSE-Adapter mainly relies on text-guided feature mixing before token injection, whereas MVFA explicitly constructs complementary text views through different aggregation strategies, uses them to guide cross-modal interaction, and then compresses the resulting multimodal information via ENQF.

In general, existing methods focus either on conventional multimodal fusion or on lightweight adaptation without sufficiently preserving complementary textual cues before multimodal compression. In contrast, our method emphasizes multi-view text-guided multimodal fusion under a frozen-LLM setting, aiming to improve cross-modal affective alignment while remaining parameter-efficient.

\section{method}

\subsection{Overall Architecture}
We propose MVFA, a parameter-efficient module that adapts frozen LLMs to MSA and ERC while preserving their general-purpose capabilities. Following standard practice in multimodal foundation model research, our approach operates on pre-extracted and standardized feature sequences rather than raw inputs such as speech waveforms, video frames, or text. This design improves reproducibility, facilitates cross-dataset transfer, and allows us to focus on the core challenges of efficient cross-modal alignment and instruction-guided affective reasoning under a frozen LLM. We adopt the MSE-Adapter preprocessing pipeline \cite{yang2025mse}, as described below.

\begin{figure*}[t]
  \vskip 0.2in
  \begin{center}
    \centerline{\includegraphics[width=\textwidth]{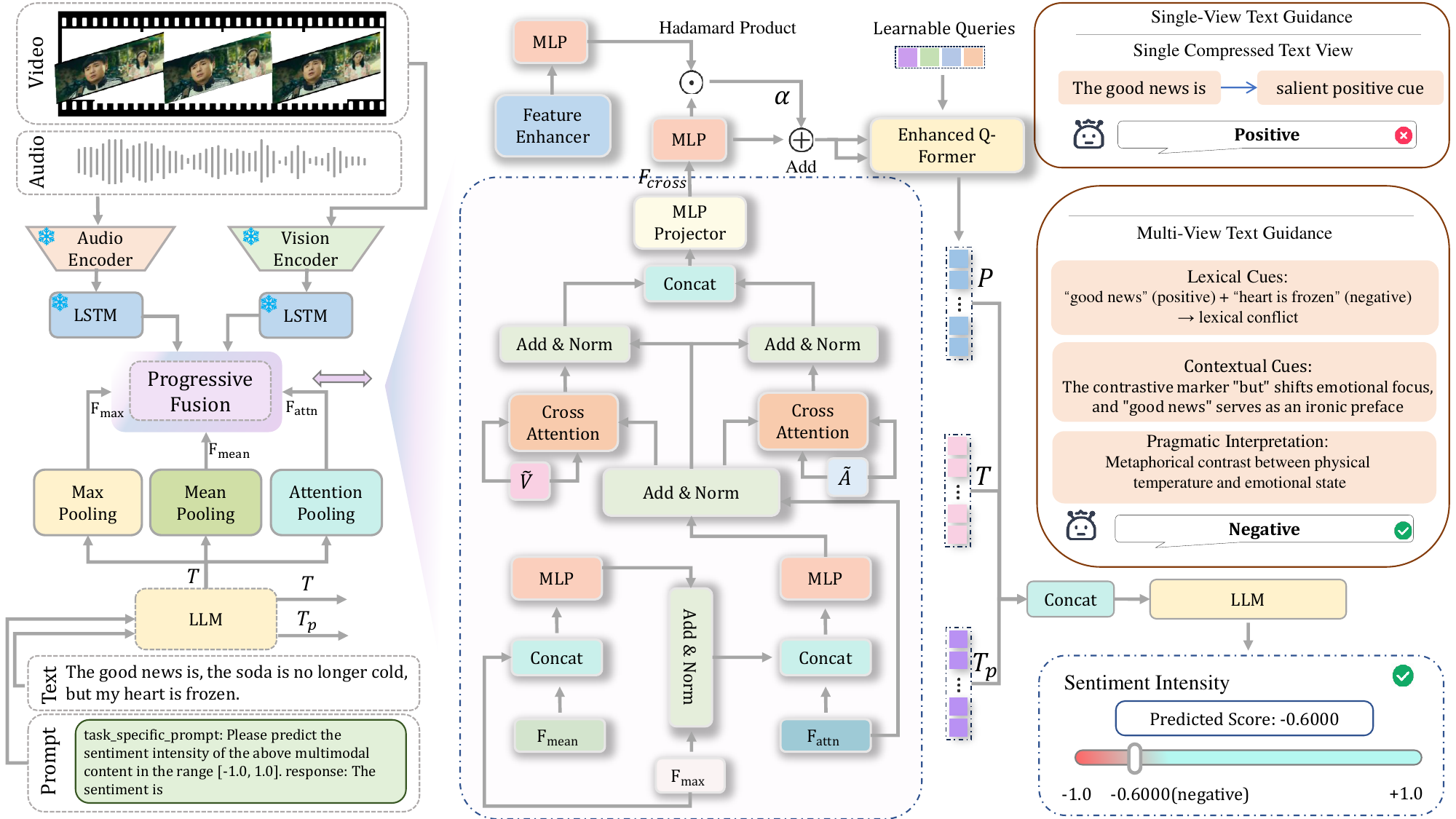}}
    \caption{Overview of the proposed MVFA framework. Lightweight encoders first extract representations from the audio and visual modalities. The MVFA then constructs complementary text views through max pooling, mean pooling, and attention pooling, progressively fuses them to guide cross-modal interaction, and compresses the resulting multimodal representation into a small set of pseudo-tokens via the Enhanced Q-Former Fusion Module (ENQF). These pseudo-tokens are concatenated with the raw text embeddings $T$ and the prompt-augmented text embeddings $T_p$ to form the input sequence $X_{\text{input}} = [P; T; T_p]$, which is fed into a frozen LLM for autoregressive sentiment or emotion prediction. Only the MVFA is trained, while the LLM backbone remains frozen.}
    \label{fig:architecture}
  \end{center}
\end{figure*}

\noindent
\textbf{Text Modality.} Each input sequence is constructed by prepending a task-specific natural-language prompt to the raw utterance. The combined sequence is tokenized using the SentencePiece tokenizer \cite{kudo2018sentencepiece} built into the target LLM, producing a sequence of token IDs. These IDs are then passed through the embedding layer of the LLM to obtain text embeddings that incorporate both task guidance and utterance semantics.

\noindent
\textbf{Audio and Visual Modalities.} For audio and visual inputs, we employ pre-trained toolkits to extract high-level semantic features \cite{liu2022make,hu2022unimse,sun2023layer}. Each toolkit produces temporal feature vectors with fixed dimensions, providing standardized representations for multimodal fusion. After preprocessing, each sample is represented by raw text embeddings $T \in \mathbb{R}^{L_t \times d_t}$, prompt-augmented text embeddings $T_p \in \mathbb{R}^{L_t \times d_t}$, audio features $A \in \mathbb{R}^{L_a \times d_a}$, and visual features $V \in \mathbb{R}^{L_v \times d_v}$, where $L_m$ and $d_m$ denote the sequence length and feature dimension of modality $m \in \{t,a,v\}$.

\subsection{Modality Encoding}

\noindent
\textbf{Audio and Visual Encoding.} For the pre-extracted temporal features $A \in \mathbb{R}^{L_a \times d_a}$ and $V \in \mathbb{R}^{L_v \times d_v}$, we employ unidirectional Long Short-Term Memory (LSTM) networks for contextual modeling. Global summaries with fixed dimensions are then obtained via linear projection:
\begin{align}
\tilde{A} &= \text{Linear}_a\left(\text{LSTM}_a(A)_{\text{final}}\right) \in \mathbb{R}^{h}, \\
\tilde{V} &= \text{Linear}_v\left(\text{LSTM}_v(V)_{\text{final}}\right) \in \mathbb{R}^{h}.
\end{align}

\noindent
\textbf{Multi-View Text Modeling and Fusion (MTF) Module.} To preserve complementary textual cues for multimodal alignment, we design a multi-view text modeling and fusion module. Let $T \in \mathbb{R}^{L_t \times d_t}$ denote the raw text embeddings, and let $h = 256$ be the intermediate representation dimension. Instead of assuming a strict semantic hierarchy, we construct three complementary text views using different aggregation strategies.

\noindent
\subsubsection{Max-Pooling Text View} An independent MLP is applied to $T$, followed by max-pooling along the sequence dimension to capture salient local activations:
\begin{equation}
F_{\text{max}} = \max(\text{MLP}_{\text{max}}(T)) \in \mathbb{R}^{h},
\end{equation}
where $\max(\cdot)$ denotes max-pooling along the sequence length $L_t$.

\noindent
\subsubsection{Mean-Pooling Text View} An independent MLP is applied to $T$, followed by average pooling along the sequence dimension to summarize the overall semantic tendency:
\begin{equation}
F_{\text{mean}} = \text{AvgPool}(\text{MLP}_{\text{mean}}(T)) \in \mathbb{R}^{h}.
\end{equation}

\noindent
\subsubsection{Attention-Pooled Text View} An independent MLP is applied to $T$, and each token is then weighted using a parameter-free attention mechanism:
\begin{gather}
S = \text{MLP}_{\text{attn}}(T) \in \mathbb{R}^{L_t \times h}, \\
a = \mathrm{softmax}(\mathrm{mean}(S, \mathrm{dim}=-1)) \in \mathbb{R}^{L_t}, \\
F_{\mathrm{attn}} = a^\top S \in \mathbb{R}^{h},
\end{gather}
where $\text{mean}(\cdot,\text{dim}=-1)$ computes a scalar score for each token by averaging over the hidden dimension.

\noindent
\subsubsection{Progressive Fusion} The three text views are fused in two stages with residual connections:
\begin{equation}
Z_1 = \text{LayerNorm}\big(\text{MLP}_1([F_{\text{max}};F_{\text{mean}}]) + F_{\text{max}}\big),
\end{equation}
\begin{equation}
F_{\text{text}} = \text{LayerNorm}\big(\text{MLP}_2([Z_1;F_{\text{attn}}]) + F_{\text{attn}}\big) \in \mathbb{R}^{h},
\end{equation}
where $[\cdot;\cdot]$ denotes concatenation. All MLP modules ($\text{MLP}_{\text{max}}$, $\text{MLP}_{\text{mean}}$, and $\text{MLP}_{\text{attn}}$) use independent parameters so that different aggregation views can learn complementary representations.

Given the fused text feature $F_{\text{text}} \in \mathbb{R}^{h}$, the global audio representation $\tilde{A} \in \mathbb{R}^{h}$, and the global visual representation $\tilde{V} \in \mathbb{R}^{h}$, we treat each vector as a sequence of length 1 and perform two independent cross-attention operations.

\noindent
\subsubsection{Text-Audio Cross-Attention} We adopt the Multi-Head Attention (MHA) mechanism of~\cite{vaswani2017attention}, using the text feature as the query and the audio feature as both key and value:
\begin{gather}
Q_a = F_{\text{text}}^\top \in \mathbb{R}^{1 \times h}, \\
K_a = V_a = \tilde{A}^\top \in \mathbb{R}^{1 \times h}, \\
O_a = \mathrm{LayerNorm}\bigl(\mathrm{MHA}(Q_a, K_a, V_a) + Q_a\bigr).
\end{gather}

\noindent
\subsubsection{Text-Visual Cross-Attention} Similarly, we use the text feature as the query and the visual feature as both key and value:
\begin{gather}
Q_v = F_{\text{text}}^\top \in \mathbb{R}^{1 \times h}, \\
K_v = V_v = \tilde{V}^\top \in \mathbb{R}^{1 \times h}, \\
O_v = \mathrm{LayerNorm}\bigl(\mathrm{MHA}(Q_v, K_v, V_v) + Q_v\bigr).
\end{gather}

Finally, the outputs of the two cross-attention operations are concatenated along the feature dimension and projected back to the original dimension through a two-layer MLP with ReLU activation:
\begin{equation}
F_{\text{cross}} = W_2 \cdot \text{ReLU}(W_1[O_a;O_v]) \in \mathbb{R}^{h},
\end{equation}
where $W_1 \in \mathbb{R}^{h \times 2h}$ and $W_2 \in \mathbb{R}^{h \times h}$ are learnable parameters.

\noindent
\textbf{Enhanced Q-Former Fusion (ENQF) Module.} To enhance cross-modal alignment while maintaining computational efficiency, we propose a streamlined Enhanced Q-Former Fusion Module. At its core is a set of learnable pseudo-queries that extract context-aware semantic representations from the fused multimodal feature via cross-attention. Formally, let $F_{\text{cross}} \in \mathbb{R}^{h}$ denote the cross-modal fused feature vector. We first project it into a normalized latent space:
\begin{equation}
F_{\text{proj}} = \mathrm{LayerNorm}\bigl(\mathrm{GELU}(W_{\text{proj}}F_{\text{cross}})\bigr),
\end{equation}
where $W_{\text{proj}}$ is a learnable projection matrix. To adaptively emphasize discriminative channels, we introduce a lightweight feature enhancement subnetwork consisting of two fully connected layers with GELU activation and a sigmoid gate:
\begin{gather}
g = \sigma(W_{g2} \cdot \text{GELU}(W_{g1}F_{\text{cross}})), \\
F_{\text{enh}} = g \odot F_{\text{proj}},
\end{gather}
where $W_{g1}$ and $W_{g2}$ are learnable parameters, $\sigma$ denotes the sigmoid function, and $\odot$ represents element-wise multiplication.
The projected feature and the enhanced feature are then fused as
\begin{equation}
F_{\text{combined}} = F_{\text{proj}} + \alpha \cdot F_{\text{enh}}.
\end{equation}
In the default configuration, we set $\alpha = 0.3$. A sensitivity analysis of $\alpha$ is provided in Sec.~\ref{sec:hyper_sensitivity}. We introduce a set of learnable query embeddings $Q \in \mathbb{R}^{n \times h}$, where the default number of pseudo-tokens is $n = 4$ and the hidden dimension is $h = 256$. After a linear transformation, cross-attention is performed with the fused feature:
\begin{equation}
H = \text{MHA}(\text{Linear}(Q), F_{\text{combined}}, F_{\text{combined}}).
\end{equation}
After residual connection and layer normalization, the output is mapped to the target dimension through a lightweight projection head to obtain the final pseudo-tokens $P$:
\begin{equation}
P = \text{Proj}_{\text{out}}(H) \in \mathbb{R}^{n \times d_t}.
\end{equation}
This design avoids the complex self-attention stacking in the traditional Q-Former. While significantly reducing the number of parameters and computational overhead, it enhances the queries' ability to capture key information from the input through an explicit feature enhancement mechanism.

\subsection{LLM Inference and Loss Function}
The pseudo-tokens, raw text embeddings, and prompt-augmented text embeddings are concatenated in sequence to form the unified final input:
\begin{equation}
X_{\text{input}} = [P; T; T_p] \in \mathbb{R}^{(n + 2L_t) \times d_t}.
\end{equation}
The frozen LLM generates the sentiment or emotion label sequence $Y = [y_1, \dots, y_m]$ in an autoregressive manner, and the training loss is:
\begin{equation}
L = -\sum_{k=1}^{m} \log p(y_k \mid X_{\text{input}}, y_{<k}; \theta_{\text{LLM}}).
\end{equation}
Gradients are only backpropagated to the MVFA module, and $\theta_{\text{LLM}}$ remains frozen.

\section{Experiments}

\subsection{Experimental Settings}
Unless otherwise specified, the main experiments are conducted on a single NVIDIA H100 GPU using Python 3.10 and PyTorch 2.2.0 with CUDA 11.8. We use ChatGLM3-6B-base~\cite{glm2024chatglm} as the primary frozen backbone model. To examine the portability of MVFA across multiple frozen LLM backbones, we additionally instantiate MVFA on LLaMA2-7B~\cite{touvron2023llama} and Qwen3-8B~\cite{qwen3technicalreport} in the main benchmark comparisons. Unless otherwise specified, ablation, hyperparameter sensitivity, and qualitative analyses are conducted on the ChatGLM3-6B-base backbone. Variants equipped with the proposed adapter are denoted with the prefix MVFA. Model optimization is performed using AdamW together with a cosine learning rate scheduler and a linear warm-up strategy. Hyperparameters are tuned separately for each dataset. To ensure stability and reproducibility, all experiments are repeated five times with different random seeds, namely 0000, 1111, 2222, 3333, and 4444. Following the protocol of~\cite{yang2025mse}, sentiment prefixes are prepended to polarity labels to reduce class ambiguity during tokenization. For the ERC task, discrete emotion categories are mapped to fixed numerical identifiers within the prompt template, which facilitates structured and consistent model generation.

\begin{table*}[!t]
  \caption{MSA results on CH-SIMS V2.0. Acc2 and F1 are computed under the non-positive/positive binarization rule, and Acc2-Weak is evaluated on the subset of samples whose ground-truth sentiment scores fall in the range [-0.4, 0.4]. Results marked with * are cited from~\cite{liu2022make}, results marked with $\dagger$ are reproduced on our experimental platform, and the remaining results are taken from the corresponding original papers.}
  \label{CH-SIMSv2}
  \centering
  \small
  \setlength{\tabcolsep}{4pt}
  \renewcommand{\arraystretch}{1.2}
  \begin{tabular*}{\textwidth}{@{\extracolsep{\fill}}lcccccc@{}}
    \toprule
    Model & Acc2$\uparrow$ & F1$\uparrow$ & Acc2-Weak$\uparrow$ & Corr$\uparrow$ & R$^2$$\uparrow$ & MAE$\downarrow$ \\
    \midrule
    TFN~\cite{zadeh2017tensor}                & 76.51 & 76.31 & 66.27 & 0.667 & 35.90 & 0.323 \\
    LMF~\cite{liu2018efficient}               & 77.05 & 77.02 & 69.34 & 0.638 & 40.64 & 0.343 \\
    MFN~\cite{zadeh2018memory}                & 75.27 & 75.24 & 66.46 & 0.606 & 32.26 & 0.355 \\
    MULT~\cite{tsai2019multimodal}            & 79.50 & 79.59 & 69.61 & 0.703 & 47.15 & 0.317 \\
    Self-MM~\cite{yu2021learning}             & 79.01 & 78.89 & 71.87 & 0.640 & 29.36 & 0.335 \\
    MISA~\cite{hazarika2020misa}              & 80.53 & 80.63 & 70.50 & 0.725 & 50.59 & 0.314 \\
    MMIM~\cite{han2021improving}              & 80.95 & 80.97 & 72.28 & 0.707 & 43.81 & 0.316 \\
    AV-MC~\cite{liu2022make}                  & 82.50 & 82.55 & 74.54 & 0.732 & \textbf{50.65} & 0.297 \\
    MSE-ChatGLM3-6B~\cite{yang2025mse}$^\dagger$ & 82.86 & 82.72 & 74.69 & 0.728 & 44.66 & 0.291 \\
    \hline
    \textbf{MVFA-LLaMA2-7B (Ours)}             & 76.48 & 76.26 & 70.60 & 0.578 & 18.99 & 0.360 \\
    \textbf{MVFA-Qwen3-8B (Ours)}              & 81.20 & 80.96 & 74.04 & 0.686 & 39.31 & 0.310 \\
    \textbf{MVFA-ChatGLM3-6B (Ours)}           & \textbf{84.62} & \textbf{84.59} & \textbf{77.27} & \textbf{0.736} & 45.34 & \textbf{0.284} \\
    \bottomrule
  \end{tabular*}
\end{table*}

\begin{table*}[!t]
\caption{Fine-grained emotion recognition results on the CHERMA dataset. We report F1 scores for each emotion category together with overall accuracy (Acc). Results marked with $\dagger$ are reproduced on our experimental platform.}
\label{tab:cherma_finegrained}
\centering
\small
\setlength{\tabcolsep}{5pt}
\renewcommand{\arraystretch}{1.2}
\begin{tabular}{lcccccccc}
\toprule
Model & Happiness$\uparrow$ & Sadness$\uparrow$ & Fear$\uparrow$ & Anger$\uparrow$ & Surprise$\uparrow$ & Disgust$\uparrow$ & Neutrality$\uparrow$ & Acc$\uparrow$ \\
\midrule
TFN~\cite{zadeh2017tensor}                      & 74.91 & 75.56 & 66.15 & 74.41 & 66.29 & 43.34 & 65.60 & 68.37 \\
LMF~\cite{liu2018efficient}                     & 74.52 & 75.83 & 66.73 & 74.55 & 65.08 & 45.70 & 65.64 & 68.23 \\
EFT~\cite{sun2023layer}                         & 74.98 & 76.88 & 67.32 & 74.85 & 66.73 & 47.48 & 64.60 & 68.72 \\
LFT~\cite{sun2023layer}                         & 75.07 & 76.29 & 66.80 & 74.88 & 66.67 & 47.74 & 65.97 & 69.05 \\
LFMIM~\cite{sun2023layer}                       & 76.60 & 77.83 & 69.44 & 75.32 & 69.83 & 50.20 & 68.24 & 70.54 \\
MULT~\cite{tsai2019multimodal}                  & 76.18 & 76.88 & 67.36 & 74.85 & 68.18 & 46.96 & 65.26 & 69.24 \\
PMR~\cite{lv2021progressive}                    & 75.68 & 76.46 & 67.97 & 75.43 & 67.37 & 48.93 & 66.59 & 69.53 \\
MSE-ChatGLM3-6B~\cite{yang2025mse}$^\dagger$     & 81.04 & 81.22 & 75.04 & 77.25 & 71.91 & 52.80 & 68.72 & 73.41 \\
\hline
\textbf{MVFA-LLaMA2-7B (Ours)}                   & 79.69 & 81.88 & 79.24 & 76.51 & 69.56 & 50.66 & 68.17 & 72.75 \\
\textbf{MVFA-Qwen3-8B (Ours)}                    & 80.30 & \textbf{82.45} & 80.87 & 76.85 & 72.21 & 52.37 & 69.49 & 73.87 \\
\textbf{MVFA-ChatGLM3-6B (Ours)}                 & \textbf{81.13} & 82.06 & \textbf{82.17} & \textbf{77.34} & \textbf{74.09} & \textbf{54.41} & \textbf{69.79} & \textbf{74.66} \\
\bottomrule
\end{tabular}
\vskip 0pt
\end{table*}

\subsection{Datasets}
To comprehensively evaluate the effectiveness and generalizability of the proposed adapter framework for MSA and ERC, experiments are conducted on three widely used datasets: CH-SIMS V2.0~\cite{liu2022make}, MELD~\cite{poria2019meld}, and CHERMA~\cite{sun2023layer}. These datasets provide multilingual coverage in English and Chinese, diverse conversational contexts, and rich multimodal signals with fine-grained emotion annotations. Such properties enable a rigorous assessment of cross-lingual and cross-task generalization performance.

\subsection{Metrics and Performance}
We evaluate MVFA-ChatGLM3-6B on standard datasets for MSA and ERC using established protocols. These evaluations include both classification and regression settings. For CH-SIMS V2.0, we report Acc2, F1, Acc2-Weak, Pearson correlation (Corr), coefficient of determination (R$^2$), and mean absolute error (MAE), and further analyze Acc3 and Acc5 in the ablation and sensitivity experiments. Following the standard evaluation protocol, Acc2 and F1 are computed under the non-positive/positive binarization rule, while Acc2-Weak is evaluated on the subset of samples whose ground-truth sentiment scores fall in the range [-0.4, 0.4], focusing on weak-sentiment recognition. For MELD, we report overall accuracy (Acc) and weighted F1 score (WF1). For CHERMA, we report overall accuracy (Acc) together with per-category F1 scores.

Table~\ref{CH-SIMSv2} compares the performance of MVFA with several baseline models on the CH-SIMS V2.0 dataset. We report both classification-oriented metrics, including Acc2, F1, and Acc2-Weak, and regression-oriented metrics, including Pearson correlation, R$^2$, and MAE, to provide a comprehensive evaluation of model performance. MVFA-ChatGLM3-6B achieves the best results on the main classification metrics, reaching 84.62\% in Acc2, 84.59\% in F1, and 77.27\% in Acc2-Weak. It also obtains the highest Pearson correlation of 0.736 and the lowest MAE of 0.284, while achieving a competitive R$^2$ value of 45.34. Compared with the reproduced MSE-ChatGLM3-6B baseline, MVFA-ChatGLM3-6B improves Acc2 by 1.76 points, F1 by 1.87 points, and Acc2-Weak by 2.58 points, while also improving Corr from 0.728 to 0.736, increasing R$^2$ from 44.66 to 45.34, and reducing MAE from 0.291 to 0.284. In addition, MVFA can be instantiated on Qwen3-8B and LLaMA2-7B, although the final performance remains backbone-dependent, with ChatGLM3-6B-base achieving the strongest overall results on CH-SIMS V2.0.

\begin{figure}[t]
    \centering
    \includegraphics[width=0.48\textwidth]{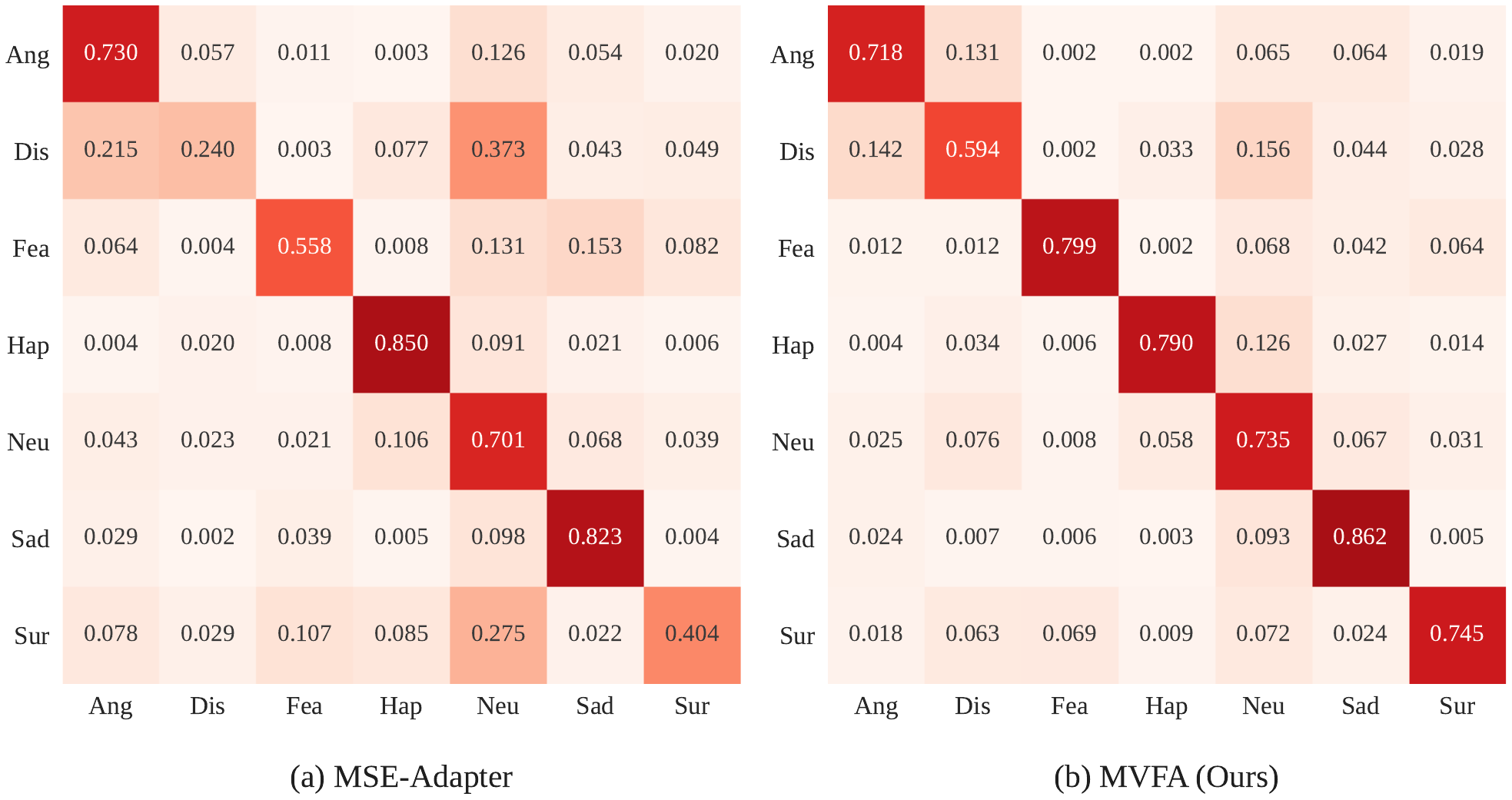}
\caption{Row-normalized confusion matrices on CHERMA for (a) MSE-Adapter and (b) MVFA, both based on ChatGLM3-6B-base. MVFA yields a more concentrated diagonal pattern, especially for difficult classes such as Disgust, Fear, and Surprise.}
\label{fig:cherma_confusion}
\end{figure}

\begin{table}[t]
\caption{Emotion recognition results on the MELD dataset. We report overall accuracy (Acc) and weighted F1 (WF1). Results marked with $\dagger$ are reproduced on our experimental platform, and the remaining results are taken from the corresponding original papers.}
\label{tab:meld_results}
\centering
\small
\setlength{\tabcolsep}{5pt}
\renewcommand{\arraystretch}{1.3}
\begin{tabular}{p{0.56\columnwidth}cc}
\toprule
Model & Acc$\uparrow$ & WF1$\uparrow$ \\
\midrule
LMF~\cite{liu2018efficient}                     & 61.15 & 58.30 \\
TFN~\cite{zadeh2017tensor}                      & 60.70 & 57.74 \\
GraphCFC~\cite{li2023graphcfc}                   & 61.42 & 58.86 \\
MM-DFN~\cite{hu2022mm}                          & 62.49 & 59.46 \\
EmoCaps~\cite{li2022emocaps}                     & --    & 64.00 \\
GA2MIF~\cite{li2023ga2mif}                       & 61.65 & 58.94 \\
UniMSE~\cite{hu2022unimse}                       & 65.09 & 65.51 \\
UniSA$_{\text{GPT2}}$~\cite{li2023unisa}         & 48.12 & 31.26 \\
UniSA$_{\text{T5}}$~\cite{li2023unisa}           & 64.52 & 62.17 \\
UniSA$_{\text{BART}}$~\cite{li2023unisa}         & 62.34 & 62.22 \\
MSE-ChatGLM3-6B~\cite{yang2025mse}$^\dagger$     & 65.88 & 64.67 \\
\hline
\textbf{MVFA-LLaMA2-7B (Ours)}                   & 65.14 & 63.66 \\
\textbf{MVFA-Qwen3-8B (Ours)}                    & 63.82 & 62.17 \\
\textbf{MVFA-ChatGLM3-6B (Ours)}                 & \textbf{67.36} & \textbf{66.03} \\
\bottomrule
\end{tabular}
\vskip 0pt
\end{table}

The fine-grained results on the CHERMA dataset are shown in Table~\ref{tab:cherma_finegrained}. In addition to the overall accuracy, we report per-category F1 scores to provide a more detailed view of model behavior across different emotion classes. Among the compared methods, MVFA-ChatGLM3-6B achieves the highest overall accuracy of 74.66\%, outperforming the reproduced MSE-ChatGLM3-6B baseline by 1.25 points and the strongest previously reported baseline, LFMIM, by 4.12 points. At the category level, MVFA-ChatGLM3-6B achieves the best F1 scores on six of the seven emotion classes, namely Happiness, Fear, Anger, Surprise, Disgust, and Neutrality, while MVFA-Qwen3-8B achieves the best result on Sadness. In addition, MVFA also remains competitive when instantiated on LLaMA2-7B and Qwen3-8B, suggesting that the proposed adapter generalizes across different frozen LLM backbones. Overall, these results indicate that the proposed multi-view text-guided fusion strategy improves not only the overall recognition performance but also fine-grained emotion discrimination across diverse affective categories.

To further examine the class-wise behavior of the proposed model, we first visualize the row-normalized confusion matrices on CHERMA for MSE-Adapter and MVFA, both instantiated on ChatGLM3-6B-base, as shown in Fig.~\ref{fig:cherma_confusion}. Compared with MSE-Adapter, MVFA exhibits a more concentrated diagonal structure overall, indicating improved fine-grained emotion discrimination. The improvement is especially evident for relatively difficult classes such as Disgust, Fear, and Surprise, whose diagonal values increase from 0.240 to 0.594, 0.558 to 0.799, and 0.404 to 0.745, respectively. MVFA also improves the recognition of Neutrality and Sadness, whereas Anger and Happiness exhibit slight decreases rather than gains. Overall, these observations suggest that MVFA is more effective at reducing confusion among several semantically adjacent emotion categories and yields more discriminative fine-grained emotion representations.

\begin{table*}[!t]
\caption{Module-level ablation results on CH-SIMS V2.0, where w/o denotes removing a module. The full model achieves the best results on the main classification-oriented metrics and on R$^2$, while some ablated variants remain competitive on individual regression-related metrics such as Corr or MAE.}
\label{tab:ablation_sims}
\centering
\small
\setlength{\tabcolsep}{4.5pt}
\renewcommand{\arraystretch}{1.2}
\begin{tabular}{lcccccccc}
\toprule
Model & Acc2$\uparrow$ & F1$\uparrow$ & Acc2-Weak$\uparrow$ & Acc3$\uparrow$ & Acc5$\uparrow$ & Corr$\uparrow$ & R$^2$$\uparrow$ & MAE$\downarrow$ \\
\midrule
w/o MTF         & 82.92 & 82.75 & 75.20 & 77.45 & 57.68 & 0.734 & 42.77 & \textbf{0.289} \\
w/o ENQF        & 82.65 & 82.46 & 75.44 & 75.90 & 55.96 & 0.710 & 43.07 & 0.296 \\
w/o MTF, ENQF   & 80.33 & 79.99 & 73.50 & 73.42 & 57.86 & \textbf{0.765} & 42.77 & 0.305 \\
\hline
\textbf{MVFA-ChatGLM3-6B (Ours)} & \textbf{84.62} & \textbf{84.59} & \textbf{77.27} & \textbf{79.01} & \textbf{59.19} & 0.736 & \textbf{45.34} & \textbf{0.284} \\
\bottomrule
\end{tabular}
\vskip 0pt
\end{table*}

\begin{table}[!t]
\caption{Module-level ablation results on MELD and CHERMA, where w/o denotes removing a module. The setting w/o MTF, ENQF corresponds to an early-fusion baseline in which features from all modalities are symmetrically concatenated with textual prompts before being fed into the frozen LLM.}
\label{tab:ablation_erc}
\centering
\small
\setlength{\tabcolsep}{5pt}
\renewcommand{\arraystretch}{1.2}
\begin{tabular}{lcccc}
\toprule
\multirow{2}{*}{Model} & \multicolumn{2}{c}{MELD} & \multicolumn{2}{c}{CHERMA} \\
\cmidrule(l){2-3} \cmidrule(r){4-5}
& Acc$\uparrow$ & WF1$\uparrow$ & Acc$\uparrow$ & WF1$\uparrow$ \\
\midrule
w/o MTF       & 66.24 & 65.04 & 73.52 & 73.21 \\
w/o ENQF      & 63.85 & 62.73 & 74.36 & 74.41 \\
w/o MTF, ENQF & 62.01 & 60.62 & 72.37 & 72.30 \\
\hline
\textbf{MVFA-ChatGLM3-6B (Ours)} & \textbf{67.36} & \textbf{66.03} & \textbf{74.66} & \textbf{74.62} \\
\bottomrule
\end{tabular}
\vskip 0pt
\end{table}

The results on the MELD dataset are reported in Table~\ref{tab:meld_results}. Among the compared methods, MVFA-ChatGLM3-6B achieves the best overall performance, reaching 67.36\% Acc and 66.03\% WF1. Compared with the reproduced MSE-ChatGLM3-6B baseline, MVFA-ChatGLM3-6B improves Acc by 1.48 points and WF1 by 1.36 points. In addition, MVFA-LLaMA2-7B achieves competitive results, while MVFA-Qwen3-8B remains weaker in this setting, suggesting that the proposed adapter is transferable across different frozen LLM backbones, although the MVFA variant built on ChatGLM3-6B-base achieves the strongest overall performance on MELD. Compared with earlier multimodal baselines such as UniMSE, MM-DFN, and GraphCFC, MVFA also shows clear improvements.

\subsection{Efficiency Analysis}

As a lightweight frozen-LLM adapter, MVFA maintains a very small trainable footprint relative to the full backbone. On the ChatGLM3-6B-base backbone, MVFA uses 6.05M--6.26M trainable parameters across the three datasets, corresponding to only 0.0968\%--0.1002\% of the full model parameters. The slight variation across datasets mainly comes from dataset-specific dimensional settings in the modality encoders, such as the LSTM-related dimensions used for audio and visual features. Overall, these results show that MVFA achieves strong multimodal adaptation while updating only about 0.1\% of the backbone parameters.

\subsection{Ablation Study}
We conduct ablation studies on three datasets, namely the English dataset MELD and the Chinese datasets CH-SIMS V2.0 and CHERMA, to evaluate the effectiveness of different components in MVFA across languages and tasks. All experimental settings, including hyperparameters and random seeds, are kept consistent with the main experiments to ensure fair comparison. For compactness, the ablation results on CHERMA are summarized using Acc and WF1, while the main results additionally report per-category F1 scores.

Table~\ref{tab:ablation_sims} reports the module-level ablation results on CH-SIMS V2.0. Removing either MTF or ENQF degrades performance, and removing both causes a further overall drop. Despite some competitive regression-related values in individual ablations, the full model achieves the best overall trade-off across metrics, indicating that both modules are beneficial and complementary.

\begin{table*}[!htbp]
\caption{Leave-one-out branch ablation of the proposed multi-view text modeling module on CH-SIMS V2.0. We remove the max-pooling, mean-pooling, and attention-pooling branches individually, denoted as w/o Max, w/o Mean, and w/o Attn.}
\label{tab:sims_multiview_ablation}
\centering
\small
\setlength{\tabcolsep}{4.5pt}
\renewcommand{\arraystretch}{1.2}
\begin{tabular}{lcccccccc}
\toprule
Model & Acc2$\uparrow$ & F1$\uparrow$ & Acc2-Weak$\uparrow$ & Acc3$\uparrow$ & Acc5$\uparrow$ & Corr$\uparrow$ & R$^2$$\uparrow$ & MAE$\downarrow$ \\
\midrule
w/o Max  & 82.92 & 82.70 & 75.40 & 77.10 & 57.37 & 0.727 & \textbf{46.32} & \textbf{0.281} \\
w/o Mean & 83.75 & 83.61 & 76.52 & 78.16 & 58.48 & 0.729 & 44.80 & 0.284 \\
w/o Attn & 82.63 & 82.33 & 75.15 & 76.42 & 56.73 & 0.728 & 45.00 & 0.286 \\
\hline
\textbf{MVFA-ChatGLM3-6B (Ours)} & \textbf{84.62} & \textbf{84.59} & \textbf{77.27} & \textbf{79.01} & \textbf{59.19} & \textbf{0.736} & 45.34 & 0.284 \\
\bottomrule
\end{tabular}
\vskip 0pt
\end{table*}

Table~\ref{tab:ablation_erc} presents the module-level ablation results on MELD and CHERMA. The full model performs best on both datasets, while removing both modules yields the largest degradation. The relative impact of MTF and ENQF varies by task, suggesting that both are important for multimodal fusion, but their contributions are dataset-dependent.

Beyond removing the whole MTF module, we further conduct leave-one-out branch ablations to examine whether the three text aggregation branches provide complementary information. Specifically, we remove the max-pooling, mean-pooling, and attention-pooling branches individually, denoted as w/o Max, w/o Mean, and w/o Attn, respectively.

Table~\ref{tab:sims_multiview_ablation} reports the leave-one-out branch ablation results on CH-SIMS V2.0. The full model achieves the best performance on the main classification-oriented metrics, while removing any single branch consistently degrades these metrics, indicating that each view contributes useful information to multimodal sentiment prediction. At the same time, some ablated variants remain competitive on individual regression-related metrics. For instance, w/o Max achieves the best MAE and the highest R$^2$, whereas the full model still attains the highest Corr and the strongest overall trade-off across metrics. These results suggest that the three branches introduce complementary inductive biases, and that their combination is more effective than any single branch alone.

We further evaluate the same multi-view design on MELD. As shown in Table~\ref{tab:meld_multiview_ablation}, the full model again achieves the best overall performance, reaching 67.36\% in accuracy and 66.03\% in weighted F1. Removing any single branch consistently degrades the results, with the largest drop observed when the mean-pooling branch is removed. This finding indicates that all three text views contribute to the final prediction, while their combination provides complementary rather than redundant information.

\begin{table}[!t]
\caption{Leave-one-out branch ablation of the proposed multi-view text modeling module on MELD. We remove the max-pooling, mean-pooling, and attention-pooling branches individually, denoted as w/o Max, w/o Mean, and w/o Attn, respectively. Results are averaged over five random seeds. $\Delta$ denotes the performance difference relative to the full MVFA model.}
\label{tab:meld_multiview_ablation}
\centering
\small
\setlength{\tabcolsep}{5pt}
\renewcommand{\arraystretch}{1.2}
\begin{tabular}{lcccc}
\toprule
Model & Acc$\uparrow$ & WF1$\uparrow$ & $\Delta$Acc & $\Delta$WF1 \\
\midrule
w/o Max  & 66.73 & 65.58 & -0.63 & -0.45 \\
w/o Mean & 66.31 & 64.89 & -1.05 & -1.14 \\
w/o Attn & 66.47 & 65.31 & -0.89 & -0.72 \\
\hline
\textbf{MVFA-ChatGLM3-6B (Ours)} & \textbf{67.36} & \textbf{66.03} & -- & -- \\
\bottomrule
\end{tabular}
\vskip 0pt
\end{table}

\begin{table}[!t]
\caption{Modality ablation results on MELD. $T$, $V$, and $A$ denote the textual, visual, and audio modalities, respectively. $\Delta$ denotes the performance difference relative to the full tri-modal model.}
\label{tab:meld_modality_ablation}
\centering
\small
\setlength{\tabcolsep}{5pt}
\renewcommand{\arraystretch}{1.2}
\begin{tabular}{lcccc}
\toprule
Modality & Acc$\uparrow$ & WF1$\uparrow$ & $\Delta$Acc & $\Delta$WF1 \\
\midrule
$V+A$   & 47.21 & 35.35 & -20.15 & -30.68 \\
$T+V$   & 66.62 & 65.62 & -0.74  & -0.41 \\
$T+A$   & 66.81 & 65.77 & -0.55  & -0.26 \\
$T+V+A$ & \textbf{67.36} & \textbf{66.03} & -- & -- \\
\bottomrule
\end{tabular}
\vskip 0pt
\end{table}

Table~\ref{tab:meld_modality_ablation} reports the modality ablation results on MELD. Here, $T$, $V$, and $A$ denote the textual, visual, and audio modalities, respectively. When the text modality is removed, the performance drops sharply: the $V + A$ setting achieves only 47.21\% Acc and 35.35\% WF1, which is substantially worse than all text-involved variants. This confirms the dominant role of textual information in ERC. At the same time, adding either visual or audio cues to text further improves performance. Specifically, both $T + A$ and $T + V$ achieve competitive results, and the difference between them is relatively small, suggesting that audio and visual modalities provide complementary non-verbal cues from different perspectives. The best performance is obtained when all three modalities are jointly fused, reaching 67.36\% Acc and 66.03\% WF1. These results demonstrate that while text serves as the primary signal, audio and visual information still provide useful complementary evidence, and MVFA can effectively exploit cross-modal interactions for ERC.

\subsection{Hyperparameter Sensitivity}
\label{sec:hyper_sensitivity}

We analyze the sensitivity of ENQF to two key hyperparameters, namely the pseudo-token number $n$ and the enhancement factor $\alpha$.

\begin{table}[t]
\caption{Sensitivity analysis of the pseudo-token number $n$ in ENQF on CH-SIMS V2.0, where $n$ denotes the number of pseudo-tokens. Results are averaged over five random seeds.}
\label{tab:pseudo_token_sensitivity}
\centering
\small
\setlength{\tabcolsep}{5pt}
\renewcommand{\arraystretch}{1.2}
\begin{tabular}{cccccccc}
\toprule
$n$ & Acc2$\uparrow$ & F1$\uparrow$ & Acc3$\uparrow$ & Acc5$\uparrow$ & Corr$\uparrow$ & R$^2$$\uparrow$ & MAE$\downarrow$ \\
\midrule
2  & 83.15 & 83.07 & 77.87 & 57.85 & 0.725 & 43.02 & 0.291 \\
4  & \textbf{84.62} & \textbf{84.59} & \textbf{79.01} & \textbf{59.19} & \textbf{0.736} & \textbf{45.34} & \textbf{0.284} \\
8  & 82.61 & 82.49 & 77.18 & 56.56 & 0.728 & 46.46 & 0.286 \\
16 & 81.20 & 80.71 & 75.57 & 55.45 & 0.725 & 45.34 & 0.291 \\
\bottomrule
\end{tabular}
\vskip 0pt
\end{table}

\begin{table}[!t]
\caption{Sensitivity analysis of the enhancement factor $\alpha$ in ENQF on CH-SIMS V2.0. Results are averaged over five random seeds, and all other settings are kept unchanged.}
\label{tab:enhancement_factor_sensitivity}
\centering
\small
\setlength{\tabcolsep}{5pt}
\renewcommand{\arraystretch}{1.2}
\begin{tabular}{cccccccc}
\toprule
$\alpha$ & Acc2$\uparrow$ & F1$\uparrow$ & Acc3$\uparrow$ & Acc5$\uparrow$ & Corr$\uparrow$ & R$^2$$\uparrow$ & MAE$\downarrow$ \\
\midrule
0.1 & 83.14 & 83.05 & 77.33 & 57.45 & 0.723 & \textbf{46.41} & \textbf{0.283} \\
0.2 & 83.66 & 83.62 & 78.41 & 59.15 & 0.727 & 44.53 & 0.285 \\
0.3 & \textbf{84.62} & \textbf{84.59} & \textbf{79.01} & \textbf{59.19} & \textbf{0.736} & 45.34 & 0.284 \\
0.4 & 83.19 & 82.97 & 77.25 & 57.00 & 0.728 & 45.67 & 0.287 \\
0.5 & 83.19 & 83.09 & 77.93 & 57.49 & 0.723 & 45.33 & 0.286 \\
\bottomrule
\end{tabular}
\vskip 0pt
\end{table}
We first study the effect of the pseudo-token number $n$, which controls the capacity of multimodal compression before token injection into the frozen LLM. As shown in Table~\ref{tab:pseudo_token_sensitivity}, we test $n \in \{2, 4, 8, 16\}$. Overall, using 4 pseudo-tokens provides the best overall trade-off among all settings. Compared with $n = 2$, setting $n = 4$ improves all the main evaluation metrics, including Acc2 (83.15\% $\rightarrow$ 84.62\%), F1 (83.07\% $\rightarrow$ 84.59\%), Acc3 (77.87\% $\rightarrow$ 79.01\%), Acc5 (57.85\% $\rightarrow$ 59.19\%), Corr (0.725 $\rightarrow$ 0.736), and R$^2$ (43.02 $\rightarrow$ 45.34), while reducing MAE from 0.291 to 0.284. Increasing the number of pseudo-tokens further to 8 or 16 does not yield consistent gains. Although $n = 8$ achieves a higher R$^2$ of 46.46, both settings underperform $n = 4$ on the main classification metrics, and $n = 16$ leads to the weakest overall results. These results suggest that too few pseudo-tokens may limit multimodal compression capacity, whereas too many may introduce redundancy and hurt overall predictive performance. We therefore use $n = 4$ in the main experiments.

We further analyze the sensitivity of the enhancement factor $\alpha$, which controls how strongly the enhanced feature contributes to the final fused representation. As shown in Table~\ref{tab:enhancement_factor_sensitivity}, we test $\alpha \in \{0.1, 0.2, 0.3, 0.4, 0.5\}$ while keeping all other settings unchanged. The results show that $\alpha=0.3$ gives the best overall performance, achieving the highest Acc2, F1, Acc3, and Corr. Smaller values such as 0.1 and 0.2 remain competitive and are slightly better on individual metrics, specifically, $\alpha=0.2$ yields the best Acc5, while $\alpha=0.1$ achieves the best R$^2$ and MAE. When $\alpha$ increases further to 0.4 or 0.5, performance degrades on most metrics. These results suggest that a moderate enhancement strength is beneficial: if the coefficient is too small, the enhancement branch is underutilized, whereas an excessively large coefficient may over-amplify auxiliary signals and disturb the primary fused representation. Therefore, we use $\alpha=0.3$ in the main experiments.

\subsection{Qualitative Case Analysis on CH-SIMS V2.0}

\begin{figure*}[t]
  \centering
\includegraphics[width=\textwidth,height=0.8\textheight,keepaspectratio]{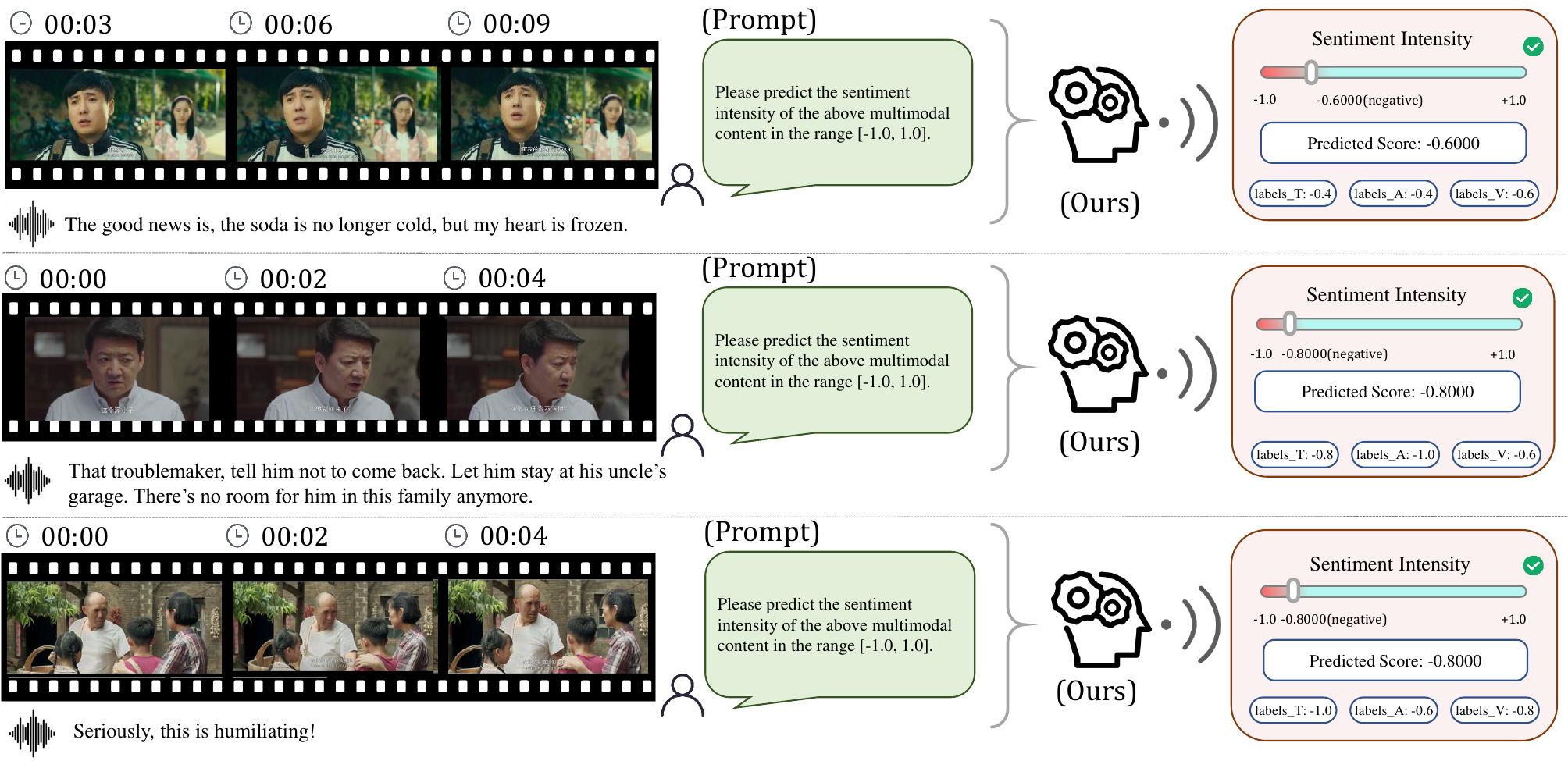}
\caption{Qualitative cases from CH-SIMS V2.0. Each example shows the utterance, predicted sentiment score, and modality-wise labels. MVFA handles both explicit and contrastive/figurative sentiment cases.}
  \label{Visualization}
\end{figure*}

To provide an intuitive view of model behavior, Fig.~\ref{Visualization} presents three qualitative cases from CH-SIMS V2.0 to illustrate how MVFA performs multimodal sentiment prediction. Each example contains the utterance, the model prediction, and the modality-wise labels for text, audio, and vision. Compared with the architectural illustration in Fig.~\ref{fig:architecture}, the purpose of this figure is not to restate the workflow, but to show how the model behaves on representative sentiment cases.

The first example is more challenging. Although the utterance contains superficially positive wording such as ``good news,'' its overall sentiment is negative due to the contrastive structure and the metaphorical expression ``my heart is frozen.'' Here, the final prediction is not determined by isolated lexical cues alone, but by the joint interpretation of textual, acoustic, and visual signals. This case qualitatively illustrates the importance of text-guided multimodal fusion: the model can move beyond literal word polarity and produce a prediction that better matches the underlying sentiment.

The second and third examples correspond to relatively explicit negative sentiment, where the textual content already carries strong affective cues and the non-textual modalities provide consistent supporting evidence. In these cases, MVFA produces strongly negative predictions that are aligned with the multimodal annotations. This suggests that the model can effectively aggregate consistent evidence across modalities when the sentiment is clearly expressed.

Overall, these examples suggest that MVFA can handle both relatively direct affective expressions and more implicit or contrastive sentiment cases. The qualitative results are consistent with the quantitative improvements reported in the main experiments, and provide an intuitive view of how multimodal evidence contributes to the final prediction.

\section{Conclusion}

We proposed MVFA, a parameter-efficient adapter for MSA and ERC with frozen LLMs. By preserving complementary text views for cross-modal interaction and compressing the fused representation into compact pseudo-tokens, MVFA enables effective multimodal reasoning without full-model fine-tuning. Experiments on CH-SIMS V2.0, MELD, and CHERMA demonstrate that MVFA achieves strong and stable performance across diverse multimodal affective datasets while remaining parameter-efficient in the frozen-LLM setting. Additional experiments on multiple frozen LLM backbones further indicate that MVFA can be applied to more than one backbone architecture, although its final performance remains dependent on the chosen backbone. These findings highlight the value of multi-view text-guided fusion for multimodal LLM adaptation. In future work, we plan to further evaluate the framework on larger backbone architectures and broader multimodal tasks.

\bibliographystyle{IEEEtran}
\bibliography{my_ref}
\end{document}